\documentclass[letterpaper, 10 pt, conference]{ieeeconf}  

\IEEEoverridecommandlockouts                              

\usepackage{graphicx}
\usepackage{amsmath}
\usepackage{amssymb}
\usepackage{caption}
\usepackage{amsmath,esint}

\usepackage{xcolor}
\usepackage[percent]{overpic}
\usepackage{float}

\usepackage{todonotes}

\newcommand{\argmin}{\operatornamewithlimits{argmin}}

\title{\LARGE \bf
Skin Normal Force Calibration Using Vacuum Bags}

\author{Joan Kangro, Silvio Traversaro, Daniele Pucci, Francesco Nori$^1$
\thanks{*This paper was supported by the FP7 EU projects CoDyCo (No. 600716 ICT 2011.2.1 Cognitive Systems and Robotics), and Koroibot (No. 611909 ICT-2013.2.1 Cognitive Systems and Robotics)}
\thanks{$^{1}$ The authors are with the iCub Facility department, Istituto Italiano di Tecnologia,
        Via Morego 30, Genoa, Italy
        {\tt\small kangrojoan@gmail.com},
        {\tt\small silvio.traversaro@iit.it},
        {\tt\small daniele.pucci@iit.it},
        {\tt\small francesco.nori@iit.it}}%
}

\begin{document}

\maketitle
\thispagestyle{empty}
\pagestyle{empty}

\begin{abstract}

The paper presents a proof of concept to calibrate iCub's skin using vacuum bags. The method's main idea consists in inserting the skin in a vacuum bag, and then decreasing the pressure in the bag 
to create a uniform pressure distribution on the skin surface. Acquisition and data processing of the bag pressure and sensors' measured capacitance allow us to characterize the relationship between  the pressure  and the measured capacitance of each sensor. After calibration, integration of the pressure distribution over the skin geometry provides us with the net normal force applied to the skin. Experiments are conducted using the forearm skin of the iCub humanoid robot, and validation results indicate 
acceptable average errors in force prediction. 

\end{abstract}

\section{INTRODUCTION}

The sense of touch is an important aspect for human-centered technology. It can be integrated to the applications such as neuroprosthetics, humanoid robotics and wearable robotics \cite{lucarotti2013synthetic}. Tactile sensors are used in order to detect the touch events and localise them. They can cover just one point or areas of larger surfaces with curvatures, edges etc. 

Tactile sensors can be roughly divided into three categories: piezoresistive, capacitive and piezoelectric. Piezoresistive sensors (e.g. \cite{kane2000traction}) are made of materials whose resistance changes with the applied force and can therefore vary the voltage of the signal. Capacitive sensors (e.g. \cite{hu2014flexible}) rely on their change in capacitance value as the dielectric between the two conductive plates is compressed. The capacitance value variation can be interpreted by the control circuit. Additionally, there are piezoelectric sensors (e.g. \cite{seminara2012tactile}) that create a voltage signal as they are deformed due to their piezoelectric properties. There are more types of tactile sensors but the mentioned are the most common ones used in the industry \cite{girao2013tactile}. 

Detecting the point of contact is important but knowing the exact force that is applied on the tactile sensors will expand the range of applications they can be used for. The papers covering tactile sensors force calibration can be roughly divided into two categories: individual sensor calibration and tactile surface calibration.

Individual sensor calibration is more common between the two. \cite{navarro2015active} focuses on calibrating the robotic hand that has pressure sensors on its fingertips. The fingers push on a plate that measures force, the data is logged and the mathematical relation between sensor value and applied force is found for each fingertip. \cite{huang2015flexible} uses similar technique for a 3D force sensor in which a known force is applied on the tactile sensor in x-y-z directions to determine the relation between sensor response and the applied force in each direction.

There is a limited number of papers that concentrate on calibrating larger areas of the skin. Two of them use a technique that involves applying mechanical force on the skin on various locations several times as the force value is modified \cite{o2015practical,maiolino2013flexible}. Neither of them give comprehensive quantitative analysis of positional nor force accuracy. Another method that was used on iCub humanoid robot uses readings from force-torque sensors, which are located in the links, and touch position information in order to find the stiffness of each of the capacitive sensor in the array \cite{ciliberto2014exploiting}. 

All the current techniques for calibrating large areas of artificial skin are time consuming. It is difficult to apply known normal force on each of the sensors in the array to find the relations between the raw values and the pressure values. However, this paper suggest a novel method to do that and trivialise the concept of artificial skin calibration. 

\begin{figure}[!t]
  \centering
  \vspace{0.4cm}
  \begin{minipage}[b]{0.23\textwidth}
    \includegraphics[width=\textwidth]{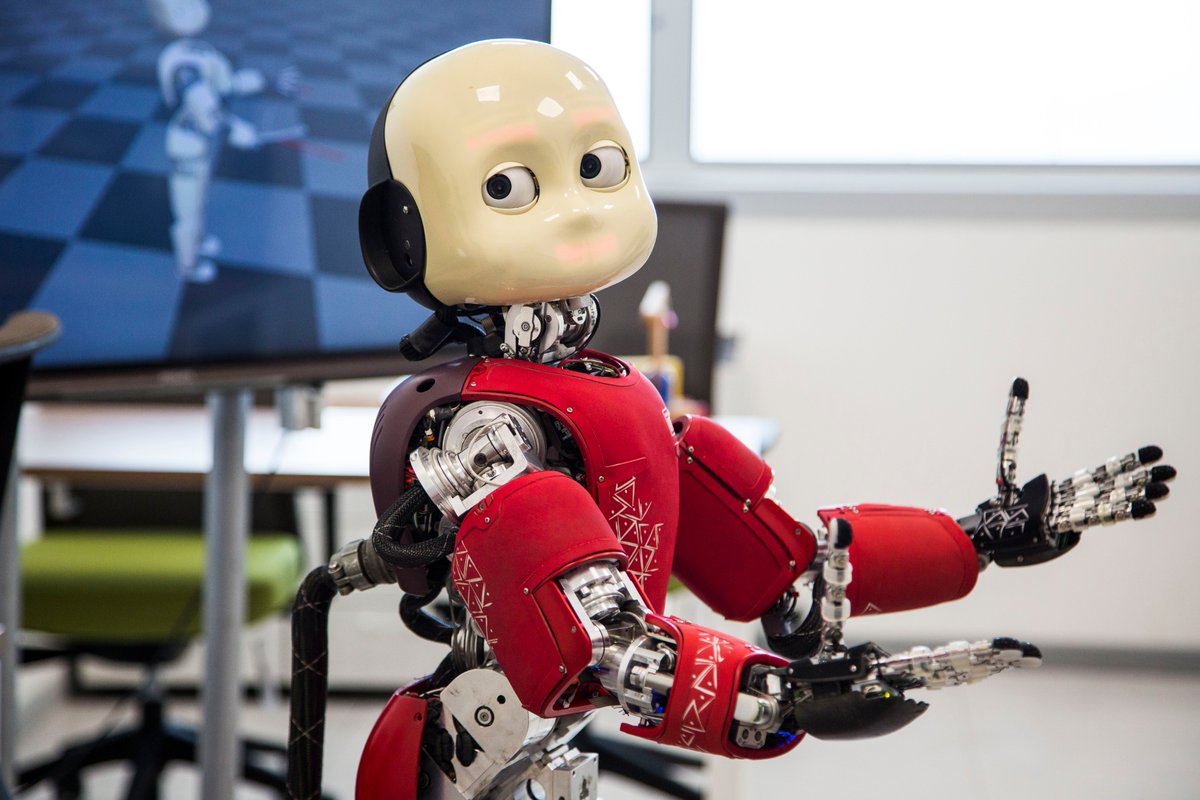}
    \caption*{(a)}
  \end{minipage}
  \hfill
  \begin{minipage}[b]{0.23\textwidth}
    \includegraphics[width=\textwidth]{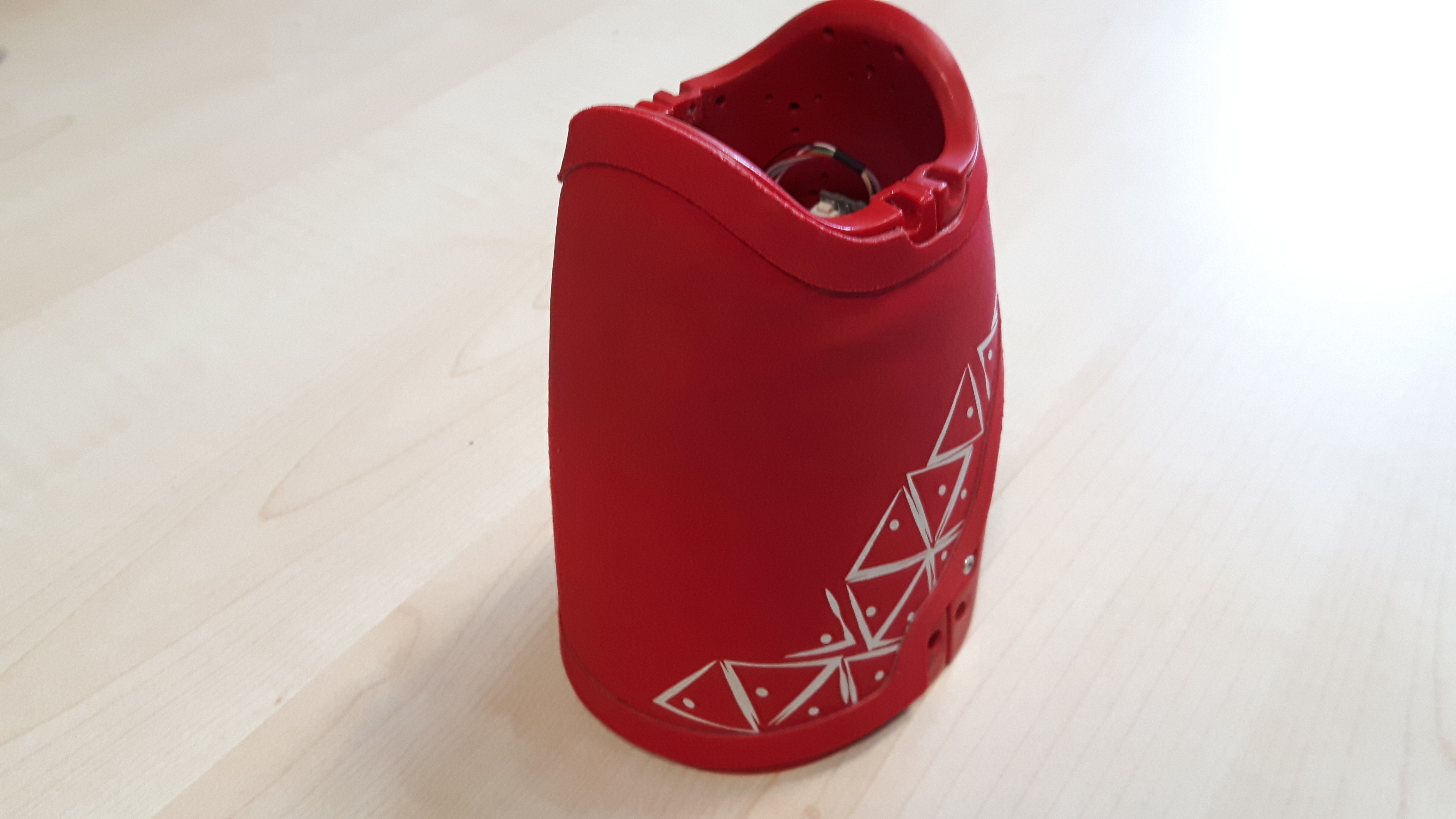}
    \caption*{(b)}
  \end{minipage}
  \caption{(a)  iCub humanoid robot;  (b) the iCub's left forearm used during the experiments.}
  \label{fig:icub_skin}
\end{figure}

This paper is organised as follows. Section II gives an overview of what kind of sensors are used during the experiments and how they work. Section III focuses on describing the calibration method in detail which includes the description of calibration experiment setup. Section IV presents the data that was extracted during the calibration experiments and what is the significance of the data. It describes how the validation experiment was conducted and the results it gave. Section V contains short summary of the paper and the main advantages of using the calibration method. It also suggests some future perspectives for increasing the accuracy of the technique.

\section{BACKGROUND}

In this paper the artificial skin under investigation is the one mounted on the left forearm of the iCub humanoid robot (Figure \ref{fig:icub_skin}). The particular skin has 230 capacitive sensors distributed over the forearm. The skin is divided into 23 triangles, each with 10 capacitive sensors (i.e. taxels) on it. The analogue signals from the sensors are converted into digital signals (using 8 bit ADC) that are sent to the computer using a CAN protocol.

\begin{figure}[t!]
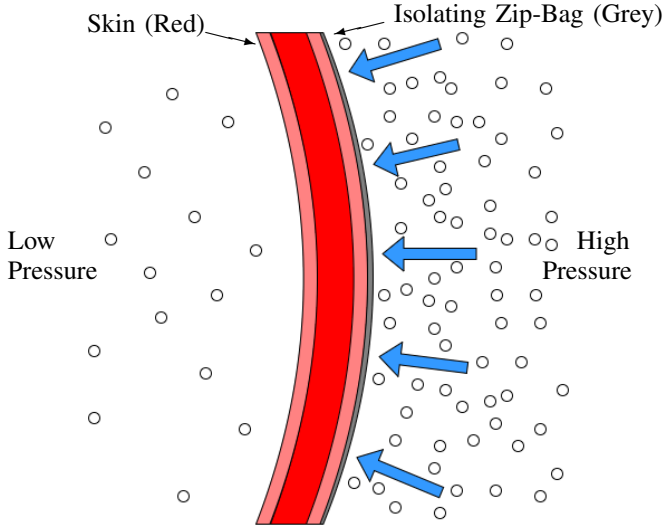

\vspace{0.8cm}
\centering
\begin{overpic}[scale=0.42]{sketch.png}
\put(-16,56){\color{black}Low}
\put(-16,50){\color{black}Pressure}
\put(99,56){\color{black}High}
\put(92,50){\color{black}Pressure}
\put(0,100){\color{black}Skin (Red)}
\put(24,100){\color{black}\vector(4,-1){10}}
\put(62,102){\color{black}Isolating Zip-Bag (Grey)}
\put(60,102){\color{black}\vector(-4,-1){10}}
\end{overpic}
\caption{
Sketch depicting how the force is induced. Light red: FPCB and ground plane; dark red:  compressed dielectric; thin grey: vacuum zip-bag; blue arrows: force distribution induced by differential pressure.}
\label{fig:sketch}
\end{figure}

The taxels consist of 3 layers. The flexible printed circuit board (FPCB) that measures the capacitance is one layer, the second layer is a soft dielectric and the third is a conductive layer that provides a common ground plane \cite{maiolino2013flexible}. Although the dielectric exhibits viscoelastic behaviour \cite{ciliberto2014exploiting}, this paper is concerned with the static behaviour of the skin and the deformation of dielectric can therefore be described by

\begin{equation} \label{eq:force_displacement}
\vec{F}=l k \hat{n}
\end{equation}
where $ \vec{F} $ represents the normal force applied on the sensor, $k$ is stiffness of the dielectric, $\hat{n}$ is the normal unit vector and $l$ is the displacement$ (|l_{initial}-l_{final}|) $. The force can also be expressed in terms of pressure using

\begin{equation} \label{eq:integration}
\vec{F} = \iint_{A} P  \hat{n} dS
\end{equation}
where $dS$ is the incremental area, $P$ is the pressure difference between the two sides of the sensor $ (|P_{outer}-P_{inner}|) $ and $A$ is the area of the sensor. This can be simplified because the differential pressure is assumed to be constant over the area of the sensor, so

\begin{equation} \label{eq:force_pressure}
\vec{F} = P A \hat{n}.
\end{equation}

As the force is applied on the sensor, the dielectric is compressed and the displacement $l$ between the FPCB and the ground plane is induced. This affects the capacitance value given by  

\begin{equation} \label{eq:capacitance_displacement}
C=\frac{\varepsilon A}{l}.
\end{equation}
where $C$ is the change in capacitance and $ \varepsilon $ is absolute permittivity of dielectric. Combining the equations (\ref{eq:force_displacement}), (\ref{eq:force_pressure}) and (\ref{eq:capacitance_displacement}) we deduce

\begin{equation} \label{eq:capacitance_pressure}
P=\frac{k \varepsilon}{C}.
\end{equation}
From this relationship it can be seen that change in capacitance value and differential pressure for the sensor is related by

\begin{equation} 
\label{eq:inverseRelation}
C \propto \frac{1}{P}.
\end{equation}

\begin{figure}
\vspace{0.8cm}
\centering
\includegraphics[scale=0.1]{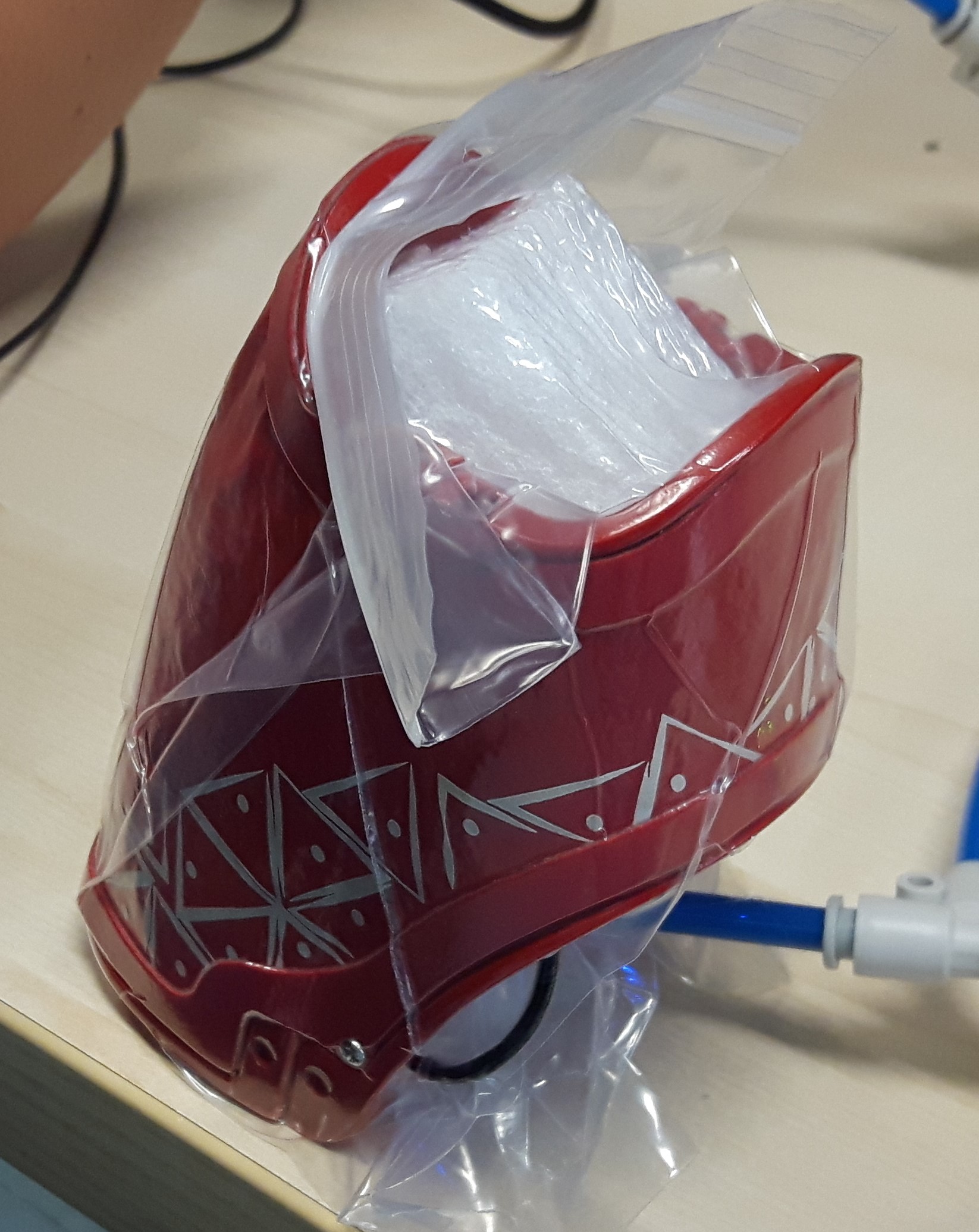}
\caption{The image displays the skin as the pressure inside the zip-bag is lowered compared to the atmospheric pressure. It can be seen that the bag is wrapped around the skin to apply uniformly distributed force over the skin.}
\label{fig:bag_closeup}
\end{figure}

However, the sensors are imperfect and in practise the returned capacitance values are dependent on a number of additional variables (curvature of the surface, thickness and stiffness of the dielectric, sensor deterioration etc.) that are complicated to model mathematically and change in time.
For this reason we model the relation between $P$ and $C$ as an arbitrary function, different for each taxel:
\begin{equation}
\label{eq:relation}
P = P(C) .
\end{equation}

\begin{figure}[t!]
\vspace{0.5cm}
\centering
\begin{overpic}[scale=0.07]{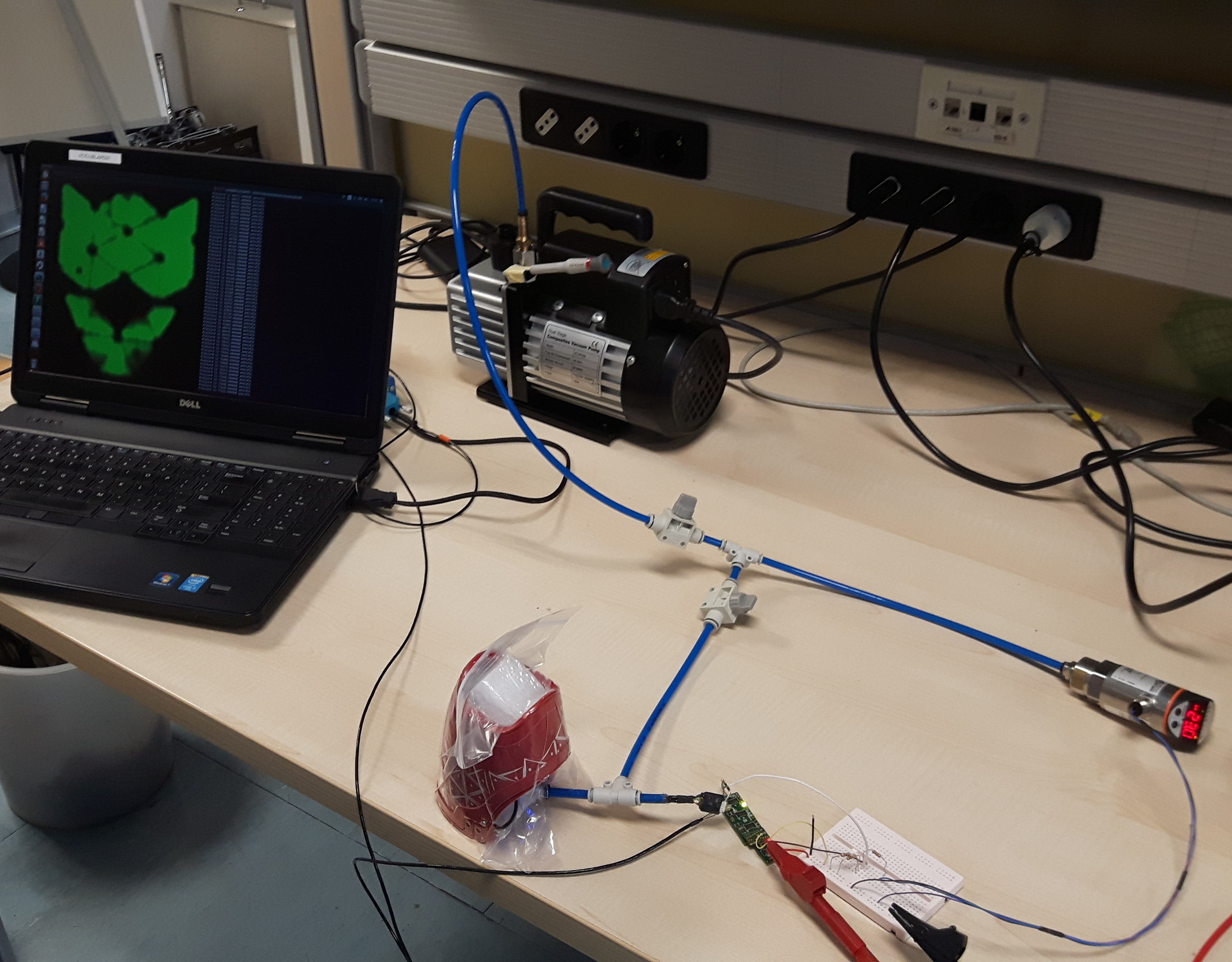}
\put(40,75){\color{white}Pump}
\put(45,73){\color{white}\vector(0,-1){10}}
\put(63,33){\color{black}Pressure Sensor}
\put(82,31){\color{black}\vector(1,-1){5}}
\put(20,19){\color{black}Skin}
\put(29,20){\color{black}\vector(1,0){6}}
\end{overpic}
\caption{Calibration setup. The vacuum pump is used to decrease the pressure in the bag, and the pressure sensor measures its pressure. The pressure sensor and the skin are interfaced to the CAN network that sends data to a PC.}
\label{fig:cal_setup}
\end{figure}

\section{CALIBRATION METHOD}
\subsection{Calibration setup}

\begin{figure}[t!]
\vspace{0.5cm}
\centering
\includegraphics[scale=0.23]{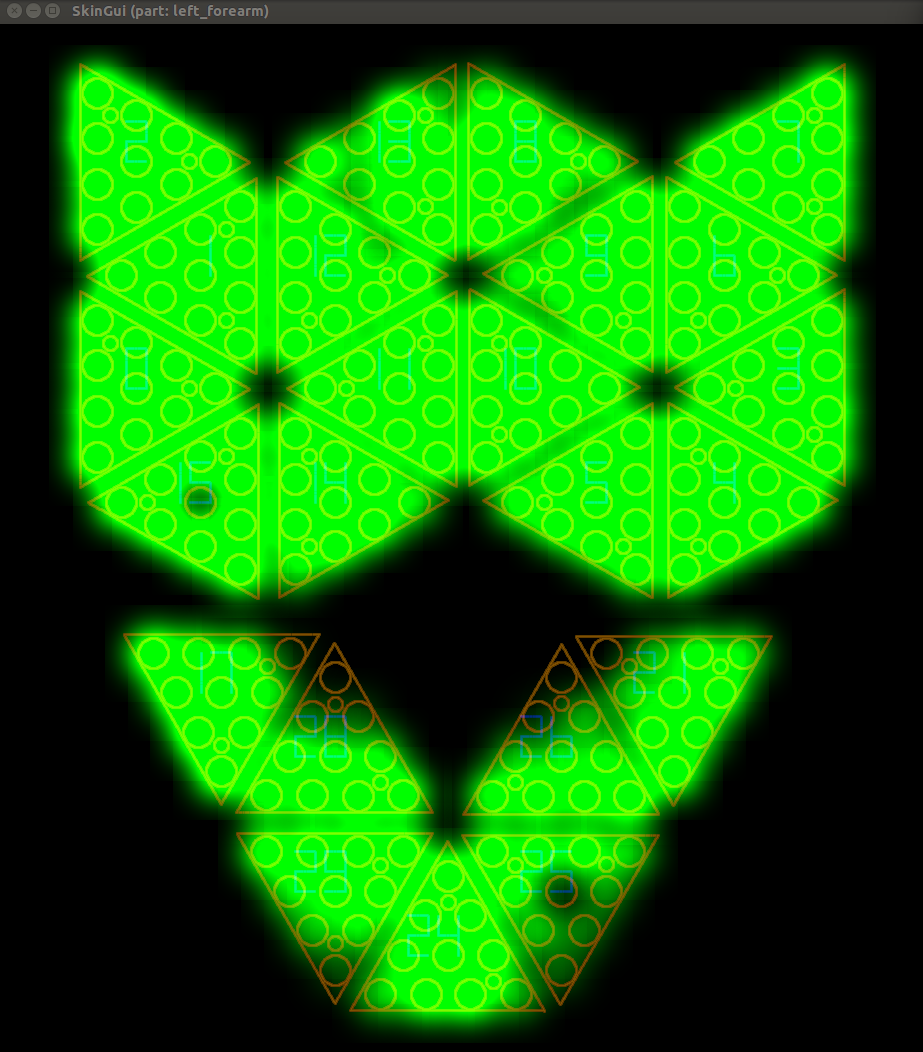}
\caption{Picture showing the user interface of the forearm skin during the experiment. The changes in capacitance values are proportional to the intensity of green. 
The variation of the green intensity indicates different responses of the sensors.}
\label{fig:calibration_UI}
\end{figure}

The calibration proposed in this paper exploits the measure of the pressure inside the bag in which the skin is placed, and from which air is pumped out. 
The image of the calibration setup is shown on figure \ref{fig:cal_setup}. The vacuum pump is used to lower the pressure inside the zip-bag compared to the atmospheric pressure to induce uniformly distributed force over the skin (see figure \ref{fig:calibration_UI}). The flowrate is regulated using the valve on the pump to ensure a slow and steady pace in order to avoid the dynamic effects of the dielectric material. Besides the zip-bag, the pump is connected to the pressure sensor using a T-connector. The differential pressure is increased throughout the experiment.

We assume that an experiment is composed by $K$ data samples, with $k = 1 \dots K$.
For each data sample $k$ we measure the $P^k$ pressure exerted on each taxel, and the raw capacitance $C_i^k$ measured by each taxel $i$.

\subsection{Model calibration}
To model \eqref{eq:relation}, i.e. the taxel-specific relation between the measured raw capacitance $C_i$ measured by taxel $i$ and the pressure $P$ exerted on the taxel, we choose a 5th order polynomial model:
\begin{equation} \label{eq:polynomial}
P(C_i) = a_{i} + b_{i} C_i + c_{i} C^2_i + d_{i} C^3_i + e_{i} C^4_i + f_{i} C^5_i
\end{equation}
where $a_{i},b_{i},c_{i},d_{i},e_{i},f_{i}$ are the taxel-specific constant representing the model for taxel $i$. 
By defining for each taxel $i$ the vector of parameters of the model as:
\begin{equation}
\pi_i = \begin{bmatrix} a_i & b_i & c_i & d_i & e_i & f_i \end{bmatrix}^\top \in \mathbb{R}^{6 \times 1} 
\end{equation}
and the \emph{regressor} of powers of capacitance $C$ as:
\begin{equation}
A(C) =  \begin{bmatrix} 1 & C & C^2 & C^3 & C^4 & C^5 \end{bmatrix} \in \mathbb{R}^{1 \times 6} .
\end{equation}
We can concisely write the polynomial model \eqref{eq:polynomial} as:
\begin{equation} \label{eq:concisePolynomial}
P(C) = A(C) \pi_i .
\end{equation}

For each taxel $i$ the estimated model parameters $\hat{\pi}_i$ can be computed from the experimental data as the solution of the following least square optimization problem: 
$$
\hat{\pi}_i = \argmin_{\pi_i} \left( \sum_{k = 1}^K \left(P^k - A(C_i^k) \pi_i\right)^2 \right)
$$

\subsection{Normal force estimation}
Once the skin pressure output is calibrated using the method illustrated in the previous subsection, we can estimate the total normal force applied on the skin.
As the force is applied on the skin after the calibration procedure, the pressures of the activated sensors (determined using the threshold value) are calculated using equation (\ref{eq:polynomial}) with the corresponding saved constants. The pressure values are multiplied by the area as given by (\ref{eq:force_pressure}) to calculate the individual forces applied on the sensors that are summed up to determine the total normal force applied on the skin. Assuming that the area of all the taxels is equal, the equation becomes
\begin{equation} \label{eq:summation}
\vec{F}_{T} = A \sum\limits_{i=1}^n  P_{i}(C_{i})  \hat{n}
\end{equation}
where $ \vec{F}_{T} $ is total normal force applied on the skin, $ A $ is the area of the taxel, $ \hat{n} $ is the normal unit vector, n is total number of activated sensors, $P_{i} $ is the pressure applied to an individual activated sensor and $ C_{i} $ is the raw capacitance value from an individual activated sensor. 

\section{EXPERIMENTAL RESULTS}

\subsection{Calibration Results}

\begin{figure}[t!]
\centering
\includegraphics[scale=0.42]{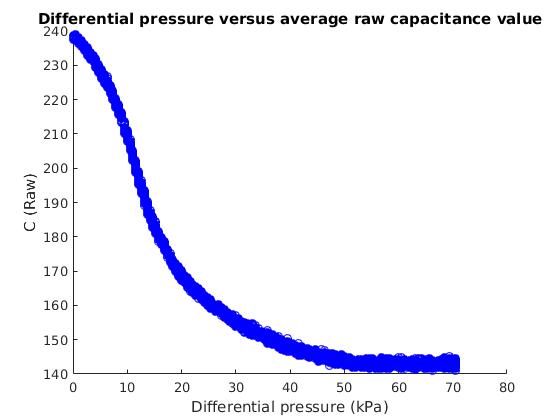}
\caption{Average sensor capacitance value versus pressure}
\label{fig:average}
\end{figure}

\begin{figure}[t!]
\centering
\includegraphics[scale=0.42]{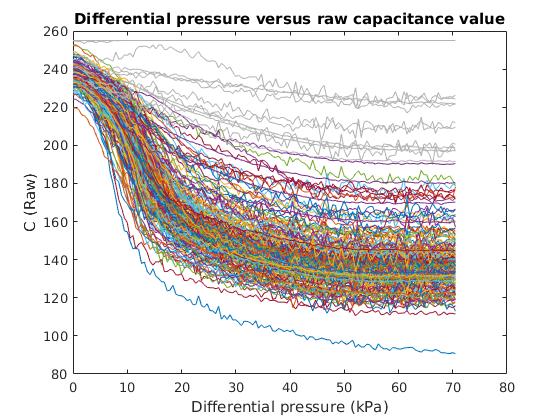}
\caption{Capacitance values (multiple values at the same pressures are averaged) of all the sensors versus pressure. The sensor responses indicated with light gray are considered inaccurate and excluded during the calibration procedure.}
\label{fig:all_sensors}
\end{figure}

The average response of all the sensors to the pressure variation (from 0 kPa to 70 kPa) is shown on figure \ref{fig:average} that has shape anticipated by the inversely proportional relation given in (\ref{eq:inverseRelation}). The curve has similar response to the one shown in \cite{maiolino2013flexible} for an individual sensor calibration (the capacitance values are inverted). Figure \ref{fig:all_sensors} shows the same relationship for all the sensors separately (multiple values at the same pressures are averaged). The critical part to point out is that some sensors have extremely low sensitivity that results in low signal to noise ratio (meaning low accuracy). One way to solve the issue is to empirically define a threshold value for the signal amplitude range ($C_{max}-C_{min}$) that determines the sensors that can be considered inaccurate (indicated with light gray on figure \ref{fig:all_sensors}). This process excludes some of the sensors that will not be used for calculating the force applied on the skin. 

\begin{figure}[!t]
  \centering
  \begin{minipage}[b]{0.42\textwidth}
    \includegraphics[width=\textwidth]{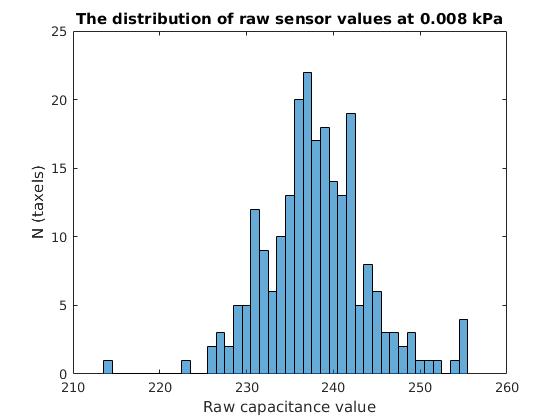}
    \caption*{(a)}
  \end{minipage}
  \hfill
  \begin{minipage}[b]{0.42\textwidth}
    \includegraphics[width=\textwidth]{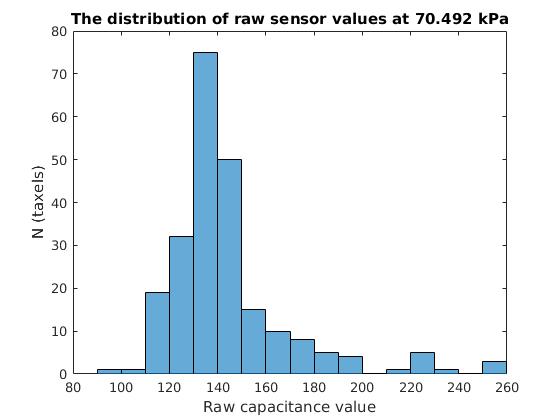}
    \caption*{(b)}
  \end{minipage}
  \caption{Distribution of the sensor capacitance values at minimum and maximum pressures. The respective standard deviations for (a) and (b) are 5.9 and 25.6 which indicates variable gains for the sensors. }
  \label{fig:sensor_distr}
\end{figure}

Additionally, figure \ref{fig:sensor_distr} indicates that sensors have variable gains as expected. This aspect is represented by the distribution of the sensor values at minimum pressure and at maximum pressure which shows that the standard deviation at maximum pressure is approximately 5 times more than at the lowest pressure. 

\begin{figure}[t!]
\centering
\includegraphics[scale=0.42]{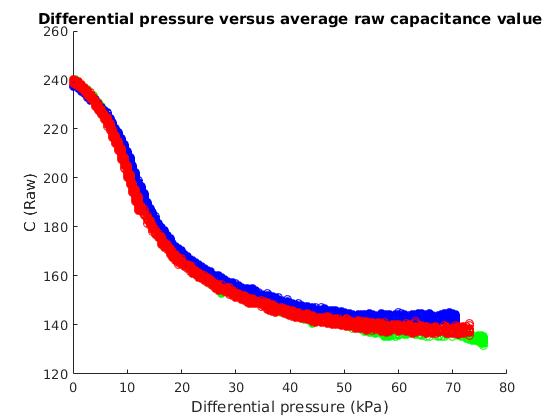}
\caption{Average sensor capacitance value versus pressure from three different experiments.}
\label{fig:multi_ave}
\end{figure}

The calibration method has relatively stable, repeatable response that is shown on figure \ref{fig:multi_ave} by plotting data from three different experiments on the same graph. It was observed that the rate of pressure change during the experiment has slight influence on the sensor response. As this paper concentrates on the static touch events, the experiments with the lowest rate of pressure change are preferred. 

\begin{figure}[t!]
\centering
\includegraphics[scale=0.42]{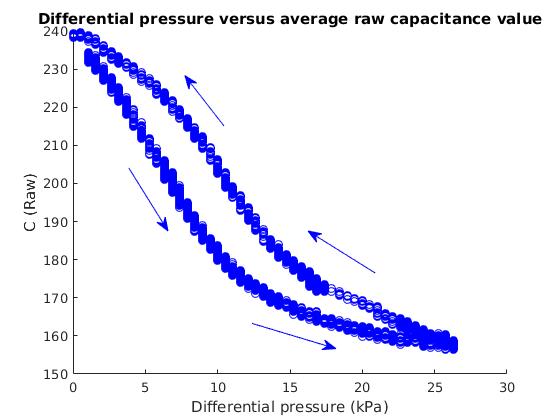}
\caption{The effect of hysteresis to average sensor response as the pressure in increased (lower curve) and then decreased (upper curve).}
\label{fig:hysteresis}
\end{figure}

The hysteresis had considerable influence during the experiments. Hysteresis effect can be observed from figure \ref{fig:hysteresis} that displays the average response of all the sensors as the pressure was increased from 0 kPa up to 26.3 kPa and then decreased back to 0 kPa. The average response of the sensors is dependent on whether the pressure is increased or decreased during the experiment. However, the calibration was performed as the pressure was increasing throughout the whole experiment to exclude the hysteresis effect. This approach was chosen because the calibration method was validated using a procedure that was not influenced by hysteresis (description in section IV(b)). 
Complete modeling and calibration of the sensor hysteresis are out of the scope of this paper.

\begin{figure}[!t]
  \centering
  \begin{minipage}[b]{0.42\textwidth}
    \includegraphics[width=\textwidth]{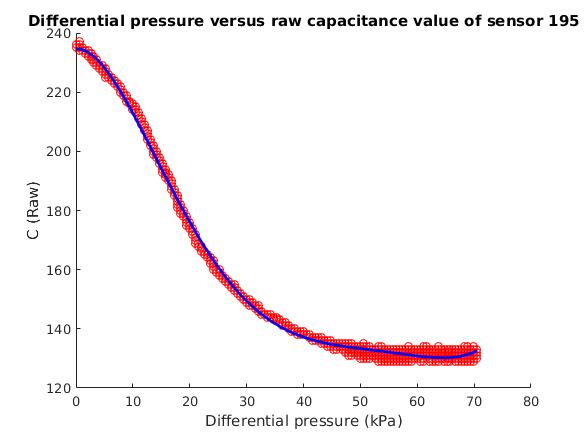}
    \caption*{(a)}
  \end{minipage}
  \hfill
  \begin{minipage}[b]{0.42\textwidth}
    \includegraphics[width=\textwidth]{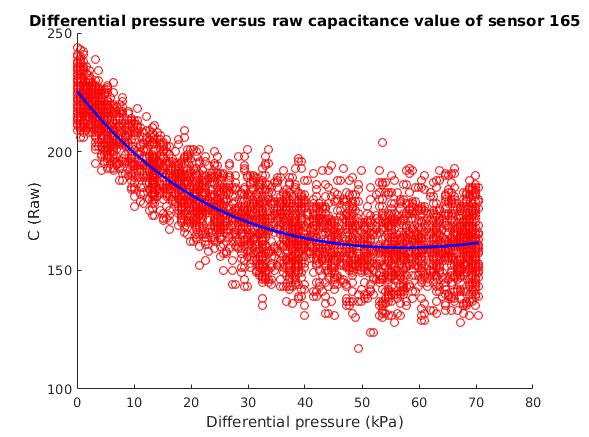}
    \caption*{(b)}
  \end{minipage}
  \caption{5th order polynomial best fit curves for sensor No. 195 (a) and sensor No. 165 (b). The graphs indicate that there are distinct responses among sensors.}
  \label{fig:fit_lines}
\end{figure}

The examples of curve fitting for two sensors are shown on figure \ref{fig:fit_lines}. The graphs illustrate that different sensors can have different gains, noise level, shape of the curve etc. The 5th order polynomial might not be the most appropriate for all the sensors but this can be optimised in the future.

\subsection{Validation}

\begin{figure}[!t]
  \vspace{0.5cm}
  \centering
  \begin{minipage}[t]{0.5\textwidth}
   \centering
    \includegraphics[width=\textwidth,angle=90]{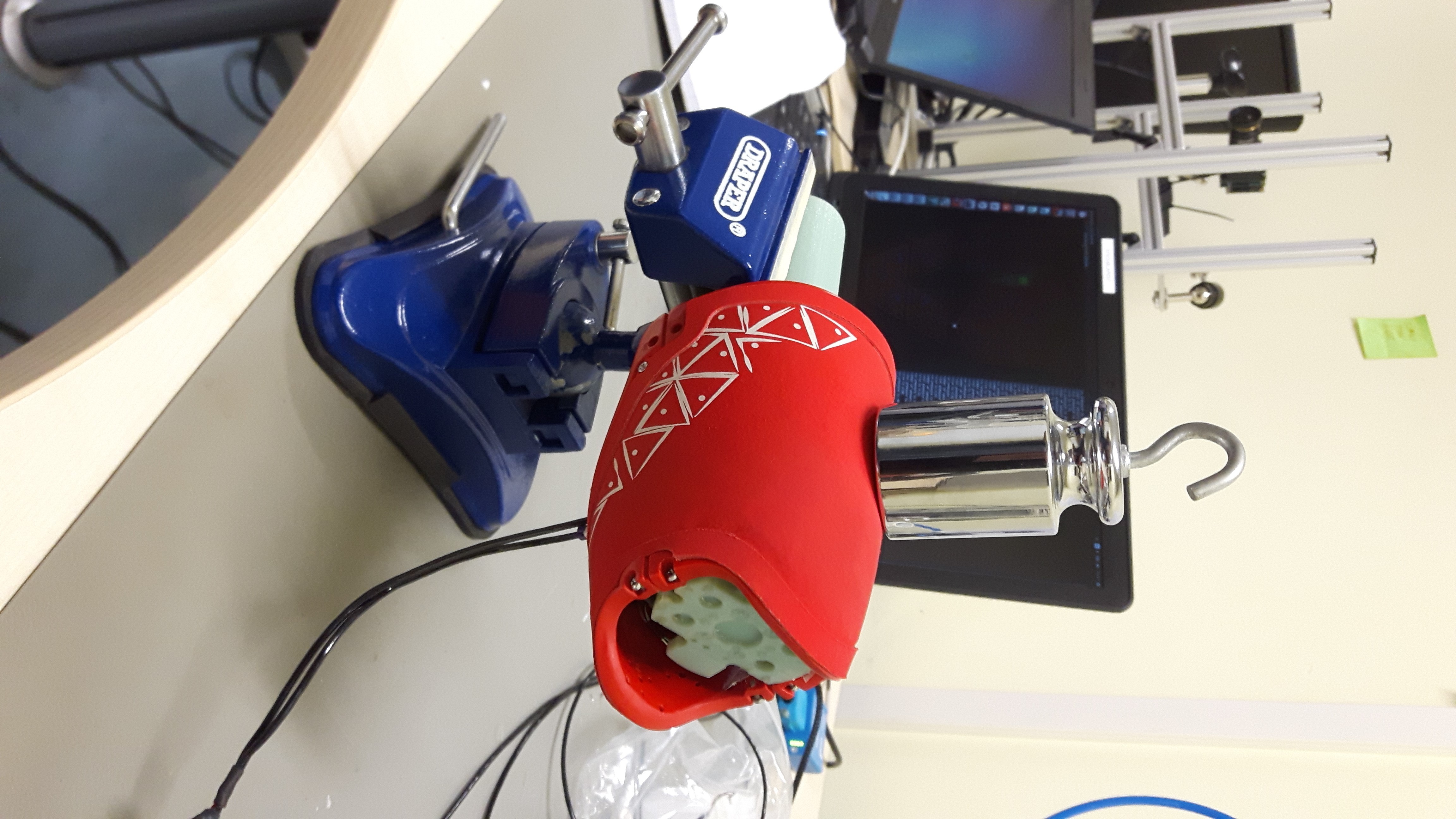}
    \caption*{(a)}
  \end{minipage}
  \hfill
  \begin{minipage}[t]{0.32\textwidth}
    \includegraphics[width=\textwidth]{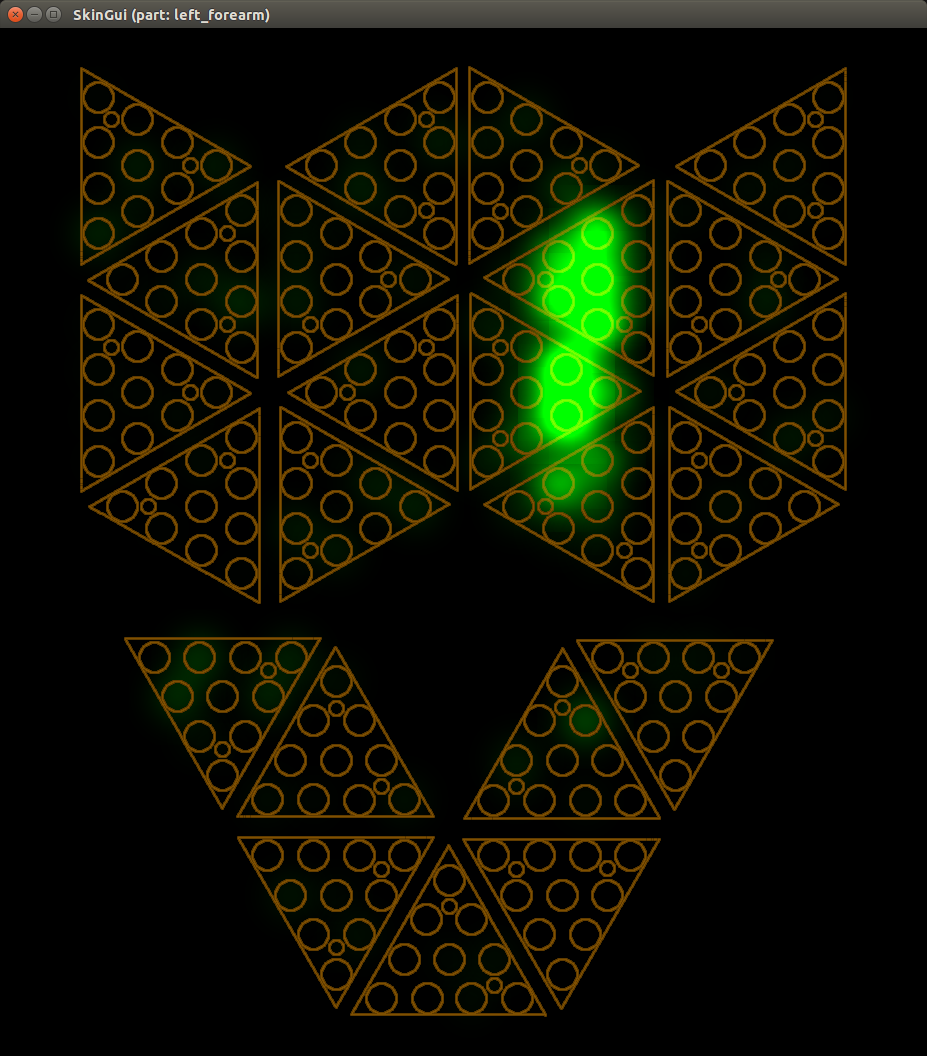}
    \caption*{(b)}
  \end{minipage}
  \caption{(a)  the validation setup. The mass is placed on top of the skin. A level is used to ensure that the skin is normal to the gravity direction. The screenshot on (b) shows the user interface of the forearm skin during the validation. The changes in capacitance values are proportional to the intensity of green color. The weight of the mass is applied on the sensors that are turned green.}
  \label{fig:validation_exp}
\end{figure}

The calibration method was validated using the weights with the known masses (range from 0.2 kg to 1 kg). They were placed on top of the skin normal to the ground as shown on figure \ref{fig:validation_exp}. The actual force applied on the body was inserted manually as an input and the program used the calibration method described in this paper to calculate the force.

\begin{figure}[t!]
\centering
\includegraphics[scale=0.42]{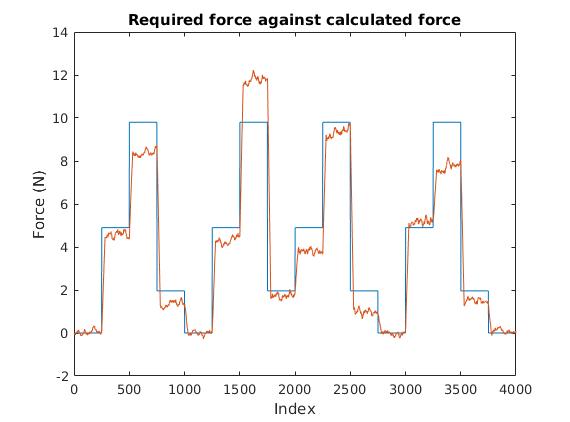}
\caption{Validation results. The calculated force is approximately following the real force applied on the body.}
\label{fig:validation}
\end{figure}

Figure \ref{fig:validation} displays the validation results. The force is calculated using the models determined during the calibration and the pressures of activated sensors are integrated over their area as given by (\ref{eq:summation}).

The force calculated is closely following the actual force applied on the skin. An average error of around 12 \% was determined during the validation.

Note that this validation error is caused by the following factors that can be reduced in the future experiments:
\begin{enumerate}
    \item {Sensor inaccuracy:} calibration method presented in this paper excludes some of the sensors that are considered inaccurate (described in section IV(a)). This creates error that is proportional to the percentage of the inaccurate sensors (roughly 5 \%).
    \item {Sensor models:} the mathematical models of the sensors are not always the most appropriate match for the training data from the calibration experiment. The improvement could be to create an algorithm that determines the order of polynomial that has the lowest value of the cost function for each particular sensor. Besides, the sensor models for validation are based on one, most stable experiment that lasted couple of minutes. The accuracy can be increased by collecting more training data.
    \item {Experimental errors:} these errors arise from the experimental setup during the calibration and validation. They include low ADC accuracy, air leaks, inclination of the masses during validation etc.
    \item {Overpressure:} the results will be inaccurate when the pressure on an individual sensor is too high and the capacitance value is over the range which the model was based on.
\end{enumerate}

\section{CONCLUSION AND DISCUSSION}

This paper presents a novel method to calibrate  artificial skin using vacuum bags. 
The main idea is to create a uniform pressure distribution over the skin surface by decreasing the air pressure in the vacuum bag. 
During the calibration, the model for the differential pressure versus capacitance value is found for each sensor. These models are used to find the pressures (and forces) applied on the skin.

The overall method results in being:
\begin{enumerate}
\item {Fast:} it takes couple of minutes to calibrate the iCub's forearm skin. 
\item {Handy:} the experiment is easy to set up and does not require any measurement of the force applied on the skin.
\item {Scalable:} it can easily be scaled to higher forces with increasing the differential pressure (instead of decreasing the pressure inside the skin, the pressure can be increased outside the isolated skin)
\item {Flexible:} the skin can have almost any shape and it does not influence the calibration results.
\end{enumerate}

The results of the calibration experiments showed that sensors have various responses that are dependent on many properties that are difficult to model mathematically. The presented calibration method can also identify sensors that are inaccurate and they will not be used during the force calculation. The results of the validation experiments demonstrated the feasibility of the method.

The planned future works related to this paper includes the following research topics:
\begin{enumerate}
    \item {Including the effect of hysteresis:} knowing the influence of hysteresis will aid with the dynamic touch force calculations.
    \item {Other skin types:} it will be useful to validate the method using other skin parts of the iCub humanoid robot and other types of artificial robotic skin. 
    \item {Creating a device:} to use the calibration technique as an everyday tool it is necessary step to create a device that can be conveniently used to calibrate various pieces of the skin. 
    \item {Reducing calibration errors:} the main errors mentioned in section IV(b) can be minimised. Most of these can be reduced with simple algorithms or better calibration set up.
\end{enumerate}

\addtolength{\textheight}{-12cm}   







\bibliographystyle{IEEEtran}  
\bibliography{main}

\end{document}